\documentclass{article} 
\usepackage{iclr2016_conference,times}
\usepackage{hyperref}
\usepackage{url}
\usepackage{times}
\usepackage{graphicx} 
\usepackage{subfig}
\usepackage{algorithmicx}
\usepackage{algpseudocode}
\usepackage{algorithm}
\usepackage{amsmath}
\usepackage{amssymb}
\usepackage{hyperref}
\usepackage{booktabs}
\usepackage[font=small,skip=5pt]{caption}

\title{Policy Distillation}

\author{Andrei A. Rusu, Sergio G\'{o}mez Colmenarejo, \c{C}a\u{g}lar G\"ul\c{c}ehre\thanks{While interning at Google DeepMind. Other affiliation: Universit\'{e} de Montr\'{e}al, Montr\'{e}al, Canada.}, Guillaume Desjardins,\\
\textbf{James Kirkpatrick, Razvan Pascanu, Volodymyr Mnih, Koray Kavukcuoglu \& Raia Hadsell} \\
Google DeepMind \\
London, UK \\
\texttt{\{andreirusu, sergomez, gdesjardins, kirkpatrick, razp, vmnih,} \\
\texttt{korayk, raia\}@google.com, gulcehrc@iro.umontreal.ca} \\
}

\begin{document}

\maketitle


\begin{abstract}
Policies for complex visual tasks have been successfully learned with
deep reinforcement learning, using an approach called deep Q-networks
(DQN), but relatively large (task-specific) networks and extensive
training are needed to achieve good performance. In this work, we
present a novel method called \emph{policy distillation} that can be
used to extract the policy of a reinforcement learning agent and train
a new network that performs at the expert level while being
dramatically smaller and more efficient. Furthermore, the same method
can be used to consolidate multiple task-specific policies into a
single policy. We
demonstrate these claims using the Atari domain and show that the
multi-task distilled agent outperforms the single-task teachers as well
as a jointly-trained DQN agent.
\end{abstract}

\section{Introduction}

Recently, advances in deep reinforcement learning have shown that policies can
be encoded through end-to-end learning from reward signals, and that these
pixel-to-action policies can deliver superhuman performance on many
challenging tasks \citep{MnihNature2015}. The deep Q-network (DQN) algorithm
interacts with an environment, receiving pixel observations and rewards. At each step, an agent chooses the action that
maximizes its predicted cumulative reward, and a convolutional network is
trained to approximate the optimal action-value function. The DQN algorithm requires long training
times to train on a single
task.


In this paper, we introduce \emph{policy distillation} for transferring one or
more action policies from Q-networks to an untrained network. The method has
multiple advantages: network size can be compressed by up to $15$ times
without degradation in performance; multiple expert policies can be combined
into a single multi-task policy that can outperform
the original experts; and finally it can be applied as a real-time,
online learning process by continually distilling the best policy
to a target network, thus efficiently tracking the
evolving Q-learning policy. The contribution of this work is to describe and
discuss the policy distillation approach and to demonstrate results on
(a) single game distillation, (b) single game distillation with highly
compressed models, (c) multi-game distillation, and (d) online distillation.

Distillation was first presented as an efficient means for supervised model
compression \citep{Caruana2006}, and it has since been extended to the problem
of creating a single network from an ensemble model
\citep{HintonDistillation2014}. It also shows merit as an optimization method
that acts to stabilize learning over large datasets or in dynamic domains \citep
{Shalev-Shwartz2014}. It uses supervised regression to train a target network to
produce the same output distribution as the original network, often using a less
peaked, or `softened' target distribution. We show that distillation can also be used in the
context of reinforcement learning (RL), a significant discovery that belies the
commonly held belief that supervised learning cannot generalize to sequential
prediction tasks \citep{barto2004dynamic}.

Distillation has been traditionally applied to networks whose outputs represent class probabilities.
In reinforcement learning, however, the neural network encodes action values, which are
real-valued and unbounded and whose scale depends on the expected future rewards in the game.
They are be blurred and non-discriminative when multiple actions have similar consequences
but sharp and discriminative when actions are important. These traits make distillation difficult to apply. 


\section{Previous Work}

This work is related to four different research areas: model compression
using distillation, deep reinforcement learning, multi-task learning and
imitation learning. The
concept of model compression through training a student network using the
outputs of a teacher network was first suggested by \citet{Caruana2006}, who
proposed it as a means of compressing a large ensemble model into a single
network. In an extension of this work, \citet{BaDistillation2013} used
compression to transfer knowledge from a deep network to a shallow network.
Other authors applied the same concept in somewhat different ways:
\citet{LiangStructureCompilation2008} proposed an approach for training a fast
logistic regression model using data labeled by a slower structured-output CRF
model; \citet{menke2009improving} used model transfer as a regularization
technique. \citet{HintonDistillation2014} introduced the term
\emph{distillation} and suggested raising the temperature of the softmax
distribution in order to transfer more knowledge from
teacher to student network. Distillation has since been applied in various ways
\citep{li2014learning, romero2014fitnets, chan2015transferring,
wang2015recurrent, tang2015knowledge}, however it has not been applied to
sequential prediction or reinforcement learning problems.


In reinforcement learning, several approaches have been proposed to learn a
policy by regression to a teacher's signal, which is often referred to as
imitation learning. Often, the teacher signal
comes from a model-based algorithm, for example in regret-based approximate policy
iteration \citep{Lazaric2010} or by using Monte Carlo tree search as an oracle
\citet{GuoSLLW2014}. In the latter case it was shown that superhuman Atari
scores could be achieved by regressing to the policy suggested by a UCT
\citep{kocsis2006bandit} algorithm. This work is related to ours, but it
requires a model of the game which has access to the true state,
rather than learning directly from observations.
The classification-based policy iteration (CAPI) framework \citep{farahmand2012}
is another approach to imitation learning which does not require a model-based
teacher. It is possbile to view a single iteration of CAPI as
policy distillation using a particular loss function (i.e. weighing classification
of actions by the action gap).  Another algorithm to tackle imitation learning is
DAGGER \cite{ross2010}. In DAGGER the student policy generates some of the training
trajectories, whereas in this work the trajectories are entirely produced by the teacher
policy.


Multi-task learning \citep{Caruana1997multitask} is often
described as a method for improving generalization performance by leveraging a
fairly limited number of \emph{similar} tasks as a shared source of inductive
bias. Typically, such tasks need to be defined on the same input distribution.
Although Atari games
share a common input modality, their images are very diverse and do not share
a common statistical basis (as opposed to natural images), making multi-task
learning much more difficult. We show that model compression and distillation
can alleviate
such issues.


\section{Approach}
\label{sec-approach}

Before describing \emph{policy distillation}, we will first give a brief review of deep Q-learning,
since DQN serves as both the baseline for performance comparisons as well as
the teacher for the policy distillation. Note that the proposed method is
not tied to DQN and can be applied to models trained using other RL
algorithms. After the DQN summary, we will describe policy distillation for
single and multiple tasks.

\subsection{Deep Q-learning}

DQN is a state-of-the-art model-free approach to reinforcement
learning using deep networks, in environments with discrete action
choices, which has achieved super-human performance on a
large collection of diverse Atari 2600 games \citep{MnihNature2015}. In
deep Q-learning, a neural network is optimized to predict the average
discounted future return of each possible action given a small number
of consecutive observations. The action with the highest predicted
return is chosen by the agent. Thus, given an environment
$\mathcal{E}$ whose interface at timestep $i$ comprises \textbf{actions} $a_i \in
\mathcal{A} = \{1,...,K\}$, \textbf{observations} $x_i \in
\mathcal{R}^d$, and \textbf{rewards} $r_i \in \mathcal{R}$, we define a
sequence $s_t = x_1, a_1, x_2, a_2, ..., a_{t-1}, x_t$ and a future
return at time $t$ with discount $\gamma$: $R_t = \sum^T_{t'=t}\gamma^{t'-t}r_t$.
The $Q$ function gives the maximum expected return after seeing sequence
$s$ and choosing action $a$: $Q^*(s,a) = \max_{\pi} \mathbb{E}[R_t|s_t=s,a_t=a,\pi]$,
where $\pi$ is an action policy, a mapping from sequences to actions. In order
to train a convolutional neural net to approximate $Q^*(s,a)$, DQN minimizes the
following loss, using samples $(s,a,r,s')$ drawn from a replay memory:

\[
L_i(\theta_i)=\mathbb{E}_{(s,a,r,s')\sim U(D)}\left[\left(r+\gamma \max_{a'} Q(s',a';\theta_i^-)-Q(s,a;\theta_i)\right)^2 \right].
\]

The use of a replay memory to decorrelate samples is a critical
element of DQN, as is the use of a \emph{target network}, an older version of
the parameters ($\theta_i^-$). Both mechanisms help to stabilize learning.

\subsection{Single-Game Policy Distillation}
\label{sec-approach-single-game}

\begin{figure}[h]
\begin{center}
  \includegraphics[width=0.8\textwidth, trim={2cm 7cm 3cm 2.5cm},clip]{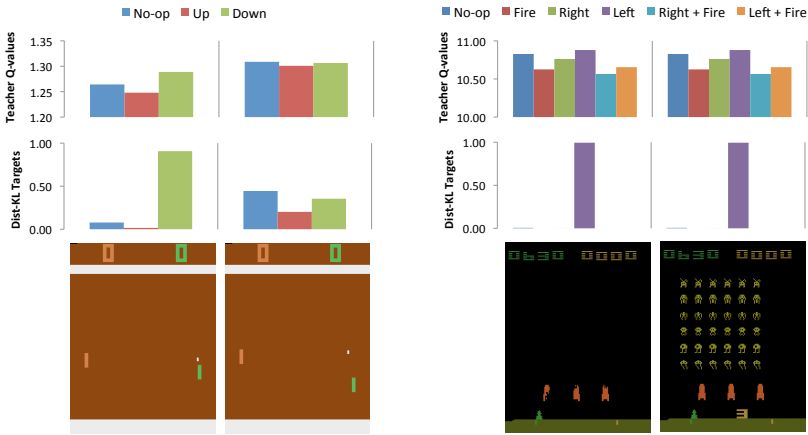}
\end{center}
\caption{Example frames from two Atari games, with the Q-values output by DQN (top) and the distillation targets after softmax (middle). For Pong, the two frames only differ by a few pixels yet the Q-values are different. In the Space Invaders example, the input frames are very different yet the Q-values are very similar. In both games the softmax sharpens the targets, making it easier for the student to learn.}
\label{pong_targets_fig}
\end{figure}

Distillation
is a method to transfer knowledge from a \emph{teacher} model $T$ to
a \emph{student} model $S$. The distillation targets from a classification
network are typically obtained by passing the weighted sums of the last network
layer through a softmax function. Figure \ref{pong_targets_fig} illustrates this with examples from two Atari games, and Figure \ref{framework_fig}(a) depicts the distillation process. In order to transfer more of the knowledge of
the network, the teacher outputs can be softened by passing the network output
through a relaxed (higher temperature) softmax than the one that was used for
training. For a selected temperature $\tau$, the new teacher outputs are thus
given by $\mathrm{softmax}(\frac{\mathbf{q}^T}{\tau})$, where $\mathbf{q}^T$ is
the vector of Q-values of $T$. These can be learned by $S$
using regression.

In the case of transferring a Q-function rather than a classifier, however,
predicting Q-values of all actions given the observation is a difficult
regression task. For one, the scale of the Q-values may be hard to learn
because it is not bounded and can be quite unstable.
Further, it is computationally challenging in general to compute the action
values of a fixed policy because it implies solving the Q-value evaluation
problem. On the other hand, training $S$ to predict only the single best
action is also problematic, since there may be multiple actions with similar
Q-values.


To test this intuition we consider three methods of policy distillation from
$T$ to $S$. In all cases we assume that the teacher $T$ has been used to
generate a dataset $\mathcal{D}^T= \{(s_i,\mathbf{q}_i)\}_{i=0}^{N}$, where
each sample consists of a short observation sequence $s_i$ and a vector
$\mathbf{q}_i$ of unnormalized Q-values with one value per action.  The first
method uses only the highest valued action from the teacher, $a_{i,best} =
\mathrm{argmax}(\mathbf{q}_i)$, and the student model is trained with a negative log
likelihood loss (NLL) to predict the same action:
\[
L_{\mathrm{NLL}}(\mathcal{D}^T,\theta_S) = -\sum_{i=1}^{|D|}\log P(a_i=a_{i,best}|x_i, \theta_S)
\]

In the second case, we train using a mean-squared-error loss (MSE). The advantage of this objective
is that it preserves the full set of action-values in the resulting student
model. In this loss, $\mathbf{q}^T$ and $\mathbf{q}^S$ are the vectors of
Q-values from the teacher and student networks respectively.
\[
L_{MSE}(\mathcal{D}^T,\theta_S) = \sum_{i=1}^{|D|} ||\mathbf{q}^T_i - \mathbf{q}^S_i||^2_2.
\]

In the third case, we adopt the distillation setup of
\citet{HintonDistillation2014} and use the Kullback-Leibler divergence (KL)
with temperature $\tau$:
\[
L_{KL}(\mathcal{D}^T,\theta_S) = \sum_{i=1}^{|D|} \mathrm{softmax}(\frac{\mathbf{q}_i^T}{\tau}) \ln \frac{\mathrm{softmax}(\frac{\mathbf{q}_i^T}{\tau})}{\mathrm{softmax}(\mathbf{q}_i^S)}
\]

In the traditional classification setting, the output distribution of
$\mathbf{q}^T$ is very peaked, so softening the distribution by raising the
temperature of the softmax allows more of the secondary knowledge to be
transferred to the student. In the case of policy distillation, however, the
outputs of the teacher are not a distribution, rather they are the expected
future discounted reward of each possible action. Rather than soften these
targets, we expect that we may need to make them sharper.

\subsection{Multi-Task Policy Distillation}
\label{multi-sec}

\begin{figure}[h]
  \centering
  \includegraphics[width=0.49\textwidth]{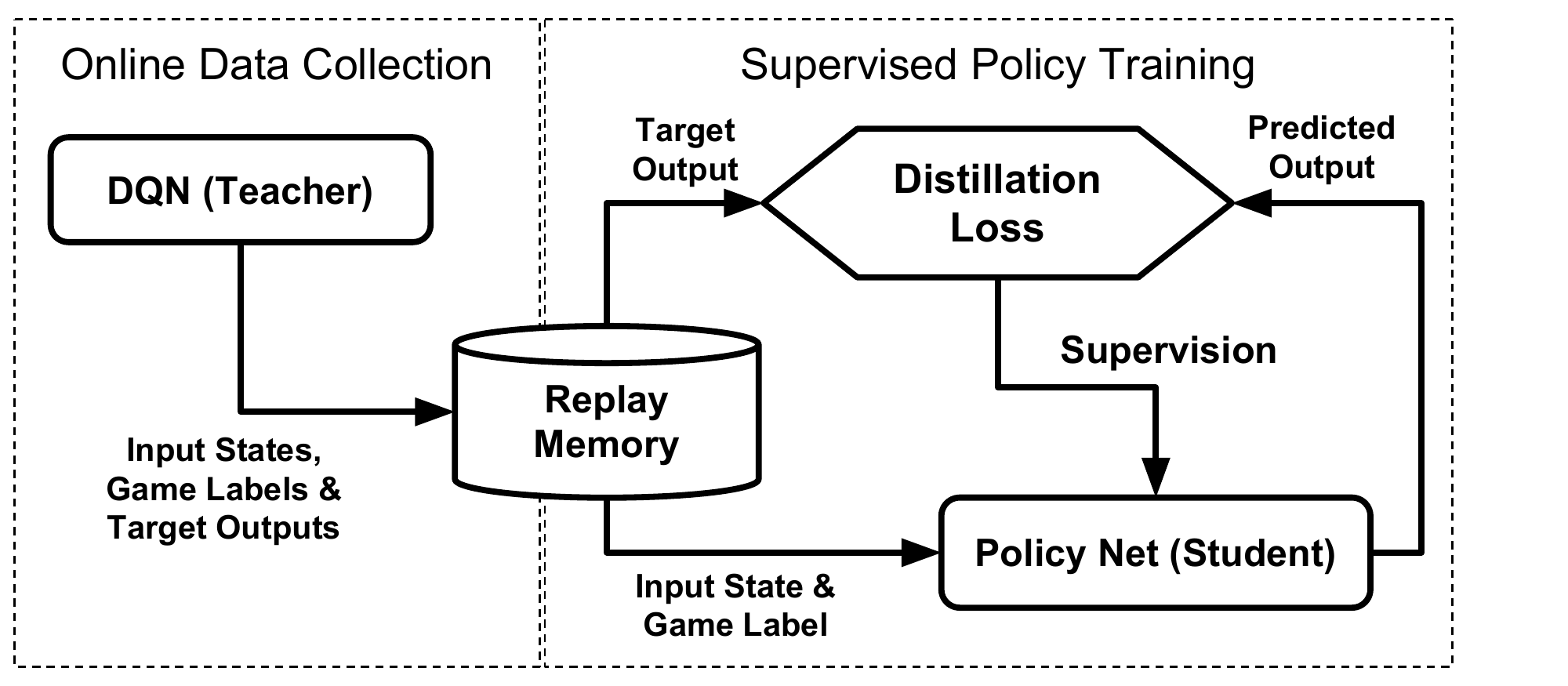}
  \includegraphics[width=0.49\textwidth]{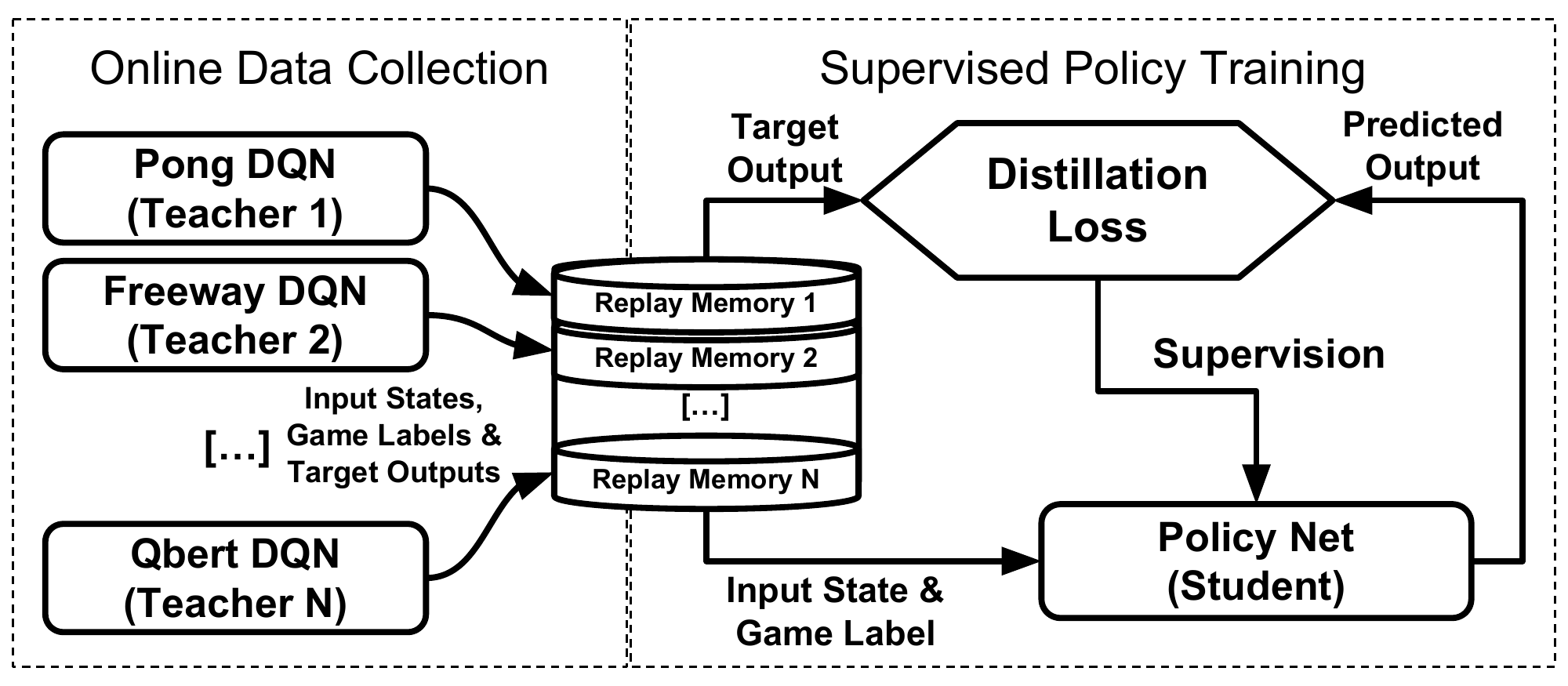}
  \caption{(a) Single-task data collection and policy distillation. The
    DQN agent periodically adds gameplay to the replay memory while
  the student network is trained. (b) Multi-task data collection and policy distillation.}
  \label{framework_fig}
\end{figure}

The approach for multi-task policy distillation, illustrated in Figure
\ref{framework_fig}(b), is straightforward. We use $n$ DQN single-game experts,
each trained separately. These agents produce inputs and targets, just as with
single-game distillation, and the data is stored in separate memory buffers. The
distillation agent then learns from the $n$ data stores sequentially, switching
to a different one every episode. Since different tasks often have different action sets, a
separate output layer (called the controller layer) is trained for each task
and the id of the task is used to switch to the correct output during both
training and evaluation. We also experiment with both the KL and NLL
distillation loss functions for multi-task learning.

An important contribution of this work is to compare the performance of multi-task
\emph{DQN} agents with multi-task \emph{distillation} agents. For multi-task DQN,
the approach is similar to single-game learning: the network is
optimized to predict the average discounted return of each possible action
given a small number of consecutive observations. Like in multi-task distillation, the game is switched every episode, separate replay memory
buffers are maintained for each task, and training is evenly interleaved
between all tasks. The game label is used to switch between
different output layers as in multi-task DQN, thus enabling a different output
layer, or controller, for each game. With this architecture in place, the multi-task DQN
loss function remains identical to single-task learning.

Even with the separate controllers, multi-game DQN learning is extremely
challenging for Atari games and DQN generally fails to reach full single-game
performance on the games. We believe this is due to interference between the
different policies, different reward scaling, and the inherent instability of
learning value functions. 

Policy distillation may offer a means of combining multiple policies into a
single network without the damaging interference and scaling problems. Since
policies are compressed and refined during the distillation process, we
surmise that they may also be more effectively combined into a single network.
Also, policies are inherently lower variance than value functions, which
should help performance and stability \citep{Greensmith2004}.




\section{Results and Discussion}

A brief overview of the training
and evaluation setup is given below; complete details are in Appendix
\ref{AppendixA}.

\subsection{Training and Evaluation}
\label{evaluation_sec}

Single task policy distillation is a process of data generation by
the \emph{teacher} network (a trained DQN agent) and supervised training by the
\emph{student} network, as illustrated in Figure \ref{framework_fig}(a). For
each game we trained a separate DQN agent, as reported in \cite{MnihNature2015}.
Each agent was subsequently fixed (no Q-learning) and used as a teacher for a
single student policy network.  The DQN teacher's outputs (Q-values
for all actions) alongside the inputs (images) were held in a buffer.
We employed a similar training procedure for multi-task policy distillation, as
shown in Figure \ref{framework_fig}(b). 

The network used to train the DQN agents is described in
\citep{MnihNature2015}. The same network was used for the student, except for
the compression experiments which scaled down the number of units in each layer.
A larger network (four times more parameters, with an additional fully connected
layer) was used to train on multi-task distillation with 10 games. The
multi-task networks had a separate MLP output (controller) layer for
each task. See Appendix\ref{AppendixA} for full details of training procedure
and networks.

Because many Atari games are highly deterministic, a learner could potentially memorize and reproduce action sequences from a few
starting points. To rigorously test the generalization capability of
both DQN teachers and distilled agents we followed the evaluation techniques
introduced by \citet{nair2015gorila} and adopted by subsequent research
\citep{vanHasselt2015double}, in which professional human expert play was used to
generate starting states for each game. Any points accumulated by the human
expert until that time were discarded, thus the agent scores are not directly
comparable to null-op evaluations previously reported. Since the agent is not in
control of the distribution over starting states, nor do we generate any
training data using human trajectories, we assert that high scores imply good
levels of generalization. 

Ten popular Atari games were selected and fixed before starting this research. These particular games were chosen in order to sample the diverse levels of DQN performance seen on the full collection, from super-human play (e.g. Breakout, Space Invaders) to below human level (Q*bert, Ms.Pacman).

\subsection{Single-Game Policy Distillation Results}

\label{comparison_of_cost_functions_subsec}

In this section we show that the Kullback-Leibler (KL) cost function
leads to the best-performing student agents, and that these distilled
agents outperform their DQN teachers on most games. Table
\ref{criteria-table} compares the effectiveness of different policy
distillation cost functions in terms of generalization performance on
four Atari games, while keeping the same network architecture as
DQN. Only four games were used for this experiment in order to
establish parameters for the loss functions which are then fixed
across other experiments (which use ten games).
Note that the evaluation uses human starting points to robustly
test generalization (see Section \ref{evaluation_sec}).

\begin{table}[h]
\centering
\caption{Comparison of learning criteria used for policy
  distillation from DQN teachers to students with identical network architectures: MSE (mean squared error),
  NLL (negative log likelihood), and KL (Kullback-Leibler divergence). Best relative scores are outlined in bold.
  }
\label{criteria-table}
\scalebox{0.9}{
\begin{tabular}{@{}l|ll|ll|ll|ll@{}}
\toprule
\textbf{} & \multicolumn{2}{l}{DQN} &
\multicolumn{2}{l}{Dist-MSE} & \multicolumn{2}{l}{Dist-NLL} &
\multicolumn{2}{l}{Dist-KL} \\ \midrule
 & \textbf{score} & & \textbf{score} & \textbf{\%DQN} & \textbf{score} & \textbf{\%DQN} & \textbf{score} & \textbf{\%DQN} \\
\textbf{Breakout} & 303.9 & & 102.9 & 33.9 & 235.9 & 77.6 & 287.8 & \textbf{94.7} \\
\textbf{Freeway} & 25.8 & & 25.7 & 99.4 & 26.2 & 101.4 & 26.7 & \textbf{103.5} \\
\textbf{Pong} & 16.2 & & 15.3 & 94.4 & 15.4 & 94.9 & 16.3 & \textbf{100.9} \\
\textbf{Q*bert} & 4589.8 & & 5607.3 & 122.2 & 6773.5 & 147.6 & 7112.8 & \textbf{155.0} \\ \bottomrule
\end{tabular}
} 
\end{table}

Students trained with a MSE loss performed worse than KL or NLL, even though
we are successfully minimizing the squared error. This is not
surprising considering that greedy action choices can be made based on
very small differences in Q-values, which receive low weight in the
MSE cost. Such situations are not uncommon in Atari games. Mean
discounted future returns are very similar in a large number of states
when coupled with control at a fine temporal resolution or very sparse
rewards. This is an intrinsic property of Q-functions, which, coupled
with residual errors of non-linear function approximation during DQN
training, make MSE a poor choice of loss function for policy
distillation.

At the other end of the spectrum, using the NLL loss assumes that a
single action choice is correct at any point in time, which is not
wrong in principle, since any optimal policy is always deterministic
if rewards are not stochastic. However, without an optimal teacher,
minimizing the NLL could amplify the noise inherent in the teacher's
learning process.

Passing Q-values through a softmax function with a temperature
parameter and minimizing the KL divergence cost strikes a convenient
balance between these two extremes. We determine empirically that a low
temperature $\tau=0.01$ is best suited for distillation in this domain. Given the performance of the KL loss, we
did not experiment with other possibilities, such as combining the NLL
and MSE criteria.

\subsection{Policy Distillation with Model Compression}

In the single game setting, we also explore model compression through
distillation. DQN networks are relatively large, in part due to optimization
problems such as local minima that are alleviated by overcomplete models. It
is also due to Q-learning, which comprises many consecutive steps of value
iteration and policy improvement, thus requiring that the same deep network must
represent a whole sequence of different policies before convergence. In
practice DQN benefits considerably from increased network capacity, but it is
likely that the final policy does not require all, or indeed, most of this
capacity.

\begin{figure}[h]
\begin{center}
\includegraphics[width=0.65\textwidth]{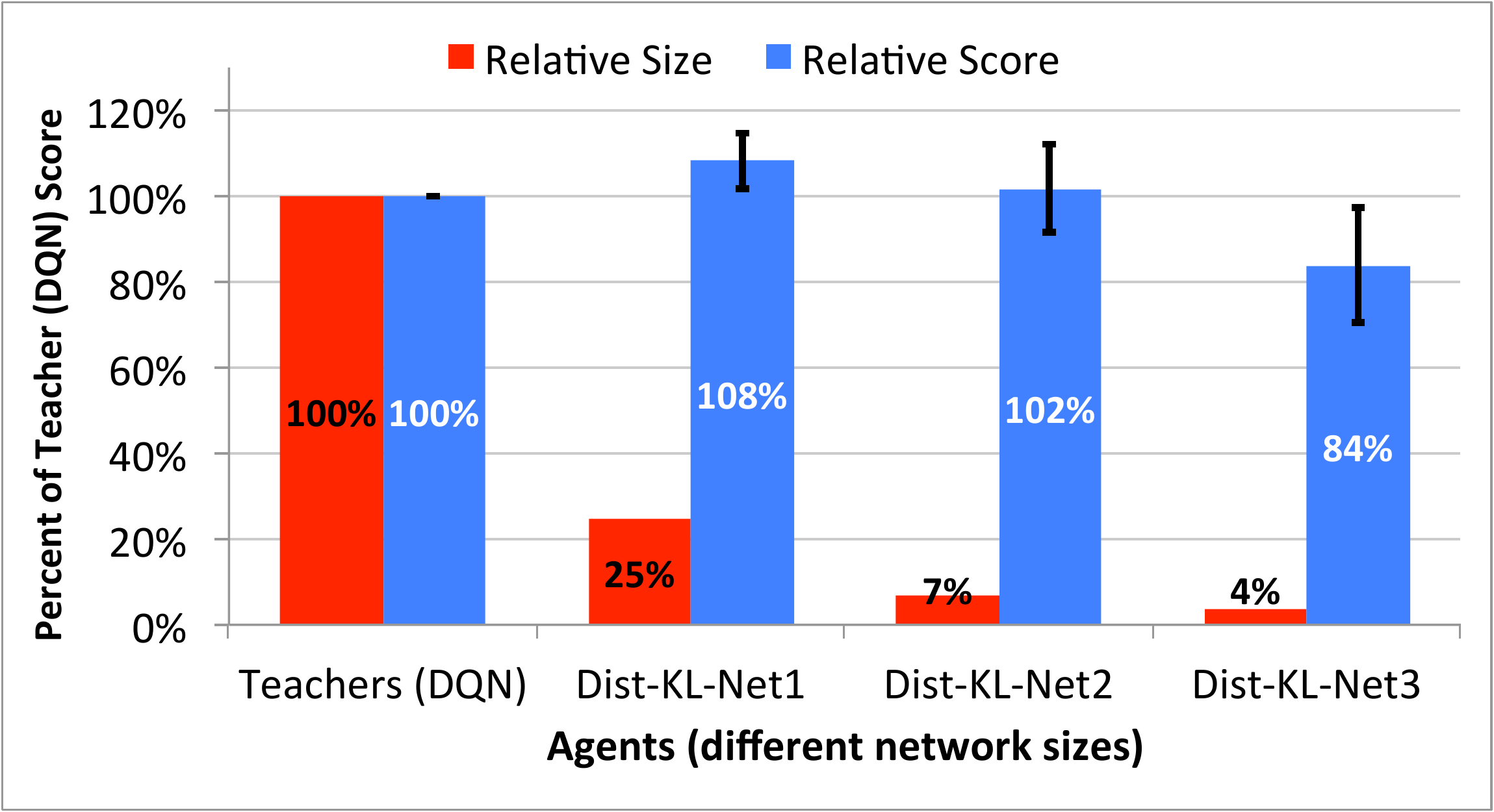}
\end{center}
\caption{Scores and sizes of distilled agents, both relative
to their respective DQN teachers. We report the geometric mean over 10 Atari games,
with error bars showing the 95\% confidence interval. A detailed results table is given in Appendix \ref{AppendixB}}
\label{single_game_fig}
\end{figure}

We evaluate single-game distilled agents and DQN teachers using 10
different Atari games, using student networks that were significantly
smaller ($25\%$, $7\%$, and $4\%$ of the DQN network parameters). The
distilled agents which are four times smaller than DQN (Dist-KL-net1,
428,000 parameters) actually outperform DQN, as shown in Figure
\ref{single_game_fig}.  Distilled agents with 15 times fewer
parameters perform on par with their DQN teachers. Even the smallest
distilled agent (Dist-KL-net3, 62,000 parameters) achieves a mean of
84\%. Details of the networks are given in Appendix \ref{AppendixA}.

These results suggest that DQN could benefit from a reduced capacity model or
regularization. However, it has been found that training a smaller DQN agent
results in considerably lower performance across games \citep{MnihNips2013}. We
speculate that training a larger network accelerates the policy iteration cycle
\citep{SuttonAndBarto1998intro} of DQN. Intuitively, once DQN performs actions
resulting in a high empirical return, it is essential that the values of the
novel trajectory are quickly estimated. A learner with limited capacity can be
very inefficient at exploiting such potential, because high returns are often
present in a minor fraction of its interactions with the environment. Hence,
strong regularization could hinder the
discovery of better policies with DQN.

\subsection{Multi-Game Policy Distillation Results}

\begin{figure}[h]
\begin{center}
  \includegraphics[width=0.65\textwidth]{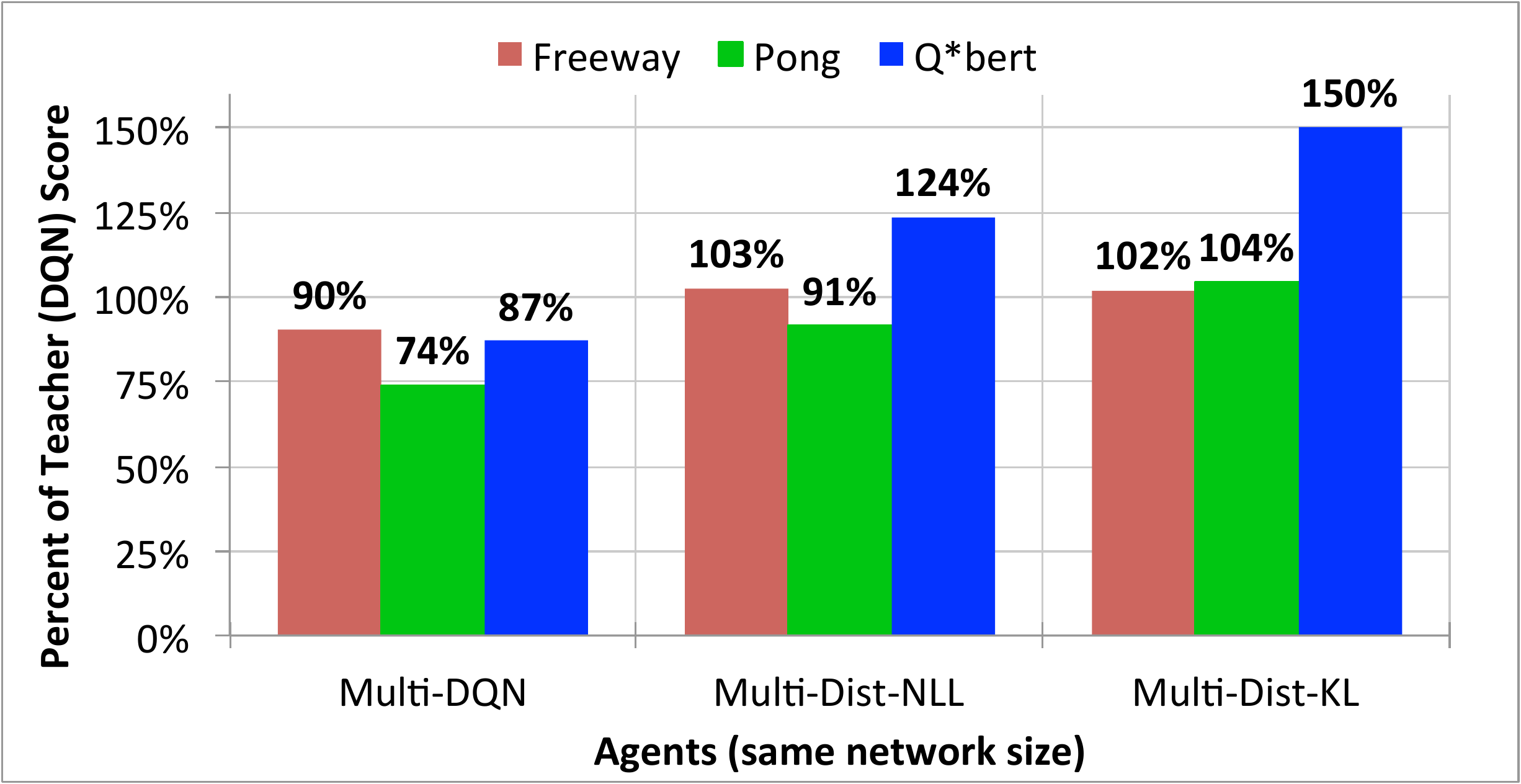}
\end{center}
\caption{Performance of multi-task agents with identical network architecture and size, relative to respective single-task DQN teachers. A detailed results table is given in Appendix \ref{AppendixB}}
\label{atari3_fig}
\end{figure}

\begin{table}[]
\centering
\caption{Performance of a distilled multi-task agent on 10 Atari games. The agent is a single network that achieves $89.3\%$ of the generalization score of 10 single-task DQN teachers, computed as a geometric mean.}
\label{atari10-table}
\scalebox{0.9}{
\begin{tabular}{@{}l|l|ll@{}}
\toprule
\textbf{}               & DQN  &
\multicolumn{2}{l}{Multi-Dist-KL} \\ \midrule
                        & \textbf{score} & \textbf{score}  &
\textbf{\% DQN}  \\
\textbf{Beamrider}      & 8672.4         & 4789.0          & 55.2
\\
\textbf{Breakout}       & 303.9          & 216.0           & 71.1
\\
\textbf{Enduro}         & 475.6          & 613.0           & 128.9
\\
\textbf{Freeway}        & 25.8           & 26.6            & 102.9
\\
\textbf{Ms.Pacman}      & 763.5          & 681.8           & 89.3
\\
\textbf{Pong}           & 16.2           & 16.1            & 99.6
\\
\textbf{Q*bert}         & 4589.8         & 6098.3          & 132.9
\\
\textbf{Seaquest}       & 2793.3         & 4320.7          & 154.7
\\
\textbf{Space Invaders} & 1449.7         & 461.1           & 31.8
\\
\textbf{Riverraid}      & 4065.3         & 4326.8          & 106.4
\\  \midrule
\textbf{Geometric Mean} &                &                 & 89.3
\\ \bottomrule
\end{tabular}
} 
\end{table}

We train a multi-task DQN agent using the standard DQN algorithm
applied to interleaved experience from three games (Multi-DQN), and
compare it against distilled agents (Multi-Dist) which were trained
using targets from three different single-game DQN teachers (see
Figure \ref{atari3_fig}). All three agents are using an identical
multi-controller architecture of comparable size to a single teacher
network. About 90\% of parameters are shared, with only 3 small MLP
``controllers'' on top which are task specific and allow for different
action sets between different games.

The multi-task DQN agent learns the three tasks to 83.5\% of single-task DQN
performance (see Figure \ref{atari3_fig}). In contrast, both distilled agents
perform better than their DQN teachers, with mean scores of 105.1\% for Multi-
Dist-NLL and 116.9\% for Multi-Dist-KL. There is a ceiling effect on Freeway and
Pong, since the single-task DQN teachers are virtually optimal, but we do see a
considerable improvement on Q*bert, with as much as 50\% higher scores for
Multi-Dist-KL.

We use the same approach to distill 10 Atari games into a single student network
that is four times larger than a single DQN. As can be seen from Table
\ref{atari10-table}, this is quite successful, with three of the games achieving
much higher scores than the teacher and an overall relative performance of
89.3\%. We don't offer a comparison to a jointly trained, 10 game DQN agent, as
was done for the three game set, because in our preliminary experiments DQN
failed to reach higher-than-chance performance on most of the games. This
highlights the challenges of multi-task reinforcement learning and supports our
findings on the three game set (Figure \ref{atari3_fig}).

\subsection{Online Policy Distillation Results}
\begin{figure*}[h]
  \centering
    \includegraphics[width=0.76\textwidth]{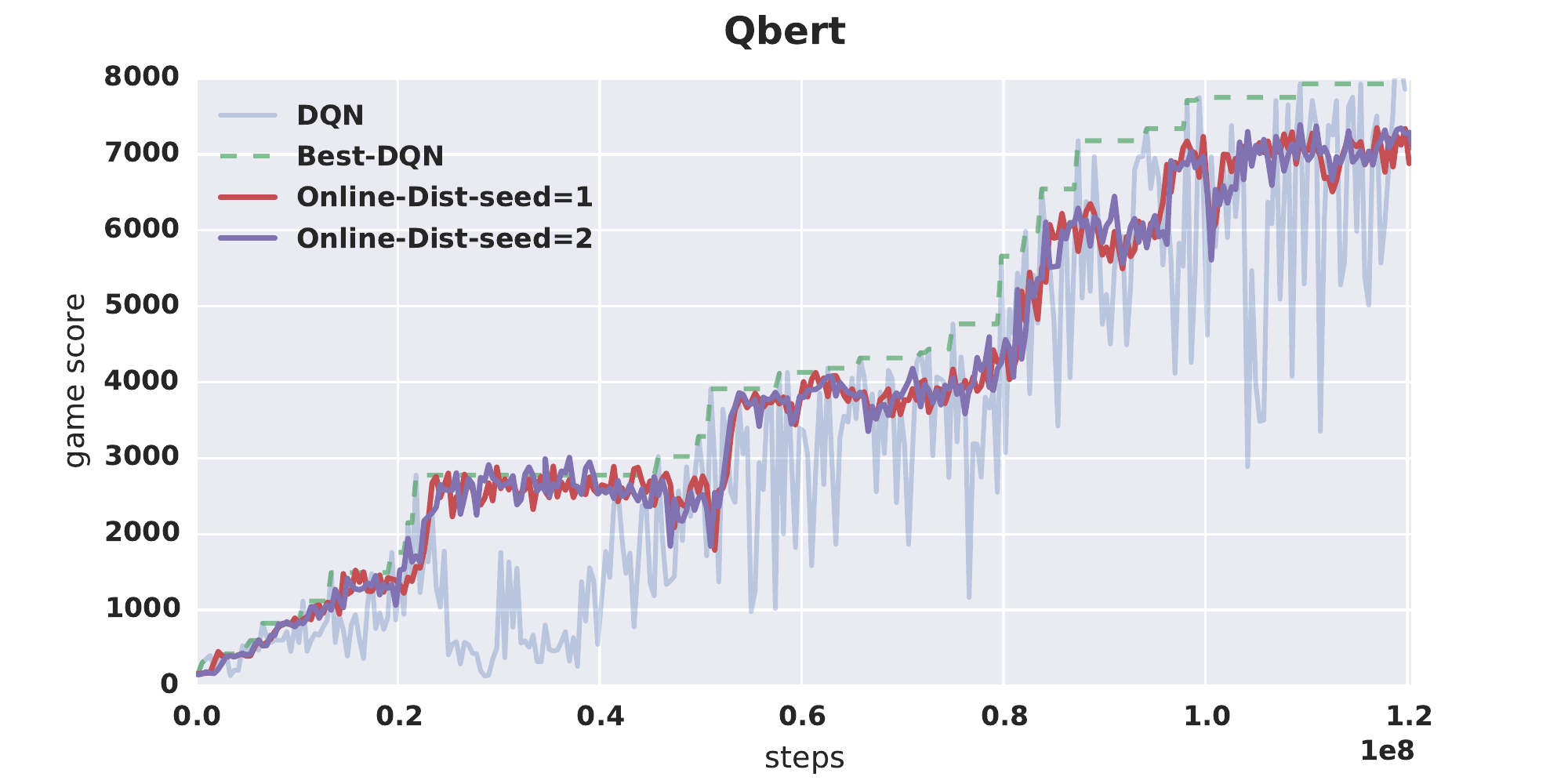}
    \caption{Online Policy Distillation during DQN learning (pale blue) on Q*bert. The current best DQN policy to date (green) is distilled into a new network  during DQN training. Showing online distillation experiments with 2 initial random seeds and the same learning rate across runs.}
  \label{online_dist_fig}
\end{figure*}

As a final contribution, we investigated online policy distillation, where the student must track the DQN teacher during Q-learning. It was not obvious whether this effort would be successful, since it has been observed that the DQN policy changes dramatically throughout training as different parts of the game are explored and mastered. To test this, DQN was trained normally and the network was periodically saved if it attained a new high score. This network was then used by the distillation learner until updated by a higher DQN score. The results of this experiment are shown in Figure \ref{online_dist_fig}. The learning curves show the high-variance DQN score, the best-so-far score, and the score reached by two distillation agents initialized with different seeds. The distilled agent is much more stable than the DQN teacher and achieves similar or equal performance on all games (see Appendix \ref{AppendixC} for additional examples of online distillation).

\section{Discussion}

In this work we have applied distillation to policy learnt in deep Q-networks.
This procedure has been used for three distinct purposes: (1) to compress
policies learnt on single games in smaller models, (2) to build agents that are
capable of playing multiple games, (3) to improve the stability of the DQN algorithm
by distilling online the policy of the best performing agent. We have shown
that in the RL setting, special care must be taken to chose the correct loss
function for distillation and have observed that the best results are obtained
by weighing action classification by a soft-max of the action-gap, similarly to
what is suggested by the CAPI framework \citet{farahmand2012}. Our results show that distillation
can be applied to reinforcement learning, even without using an iterative approach and
without allowing the student network to control the data distribution it is trained on.
The fact that the distilled policy can yield better results than the teacher
confirms the growing body of evidence that distillation is a general principle
for model regularization.

\small{\bibliography{current}}

\begin{thebibliography}{27}
\providecommand{\natexlab}[1]{#1}
\providecommand{\url}[1]{\texttt{#1}}
\expandafter\ifx\csname urlstyle\endcsname\relax
  \providecommand{\doi}[1]{doi: #1}\else
  \providecommand{\doi}{doi: \begingroup \urlstyle{rm}\Url}\fi

\bibitem[Ba and Caruana(2014)]{BaDistillation2013}
Jimmy Ba and Rich Caruana.
\newblock Do deep nets really need to be deep?
\newblock In \emph{Advances in Neural Information Processing Systems (NIPS)},
  pages 2654--2662. Curran Associates, Inc., 2014.

\bibitem[Barto and Dietterich(2004)]{barto2004dynamic}
A.~G. Barto and T.~G. Dietterich.
\newblock Handbook of learning and approximate dynamic programming.
\newblock Wiley-IEEE Press, 2004.

\bibitem[Bucila et~al.(2006)Bucila, Caruana, and Niculescu-Mizil]{Caruana2006}
Cristian Bucila, Rich Caruana, and Alexandru Niculescu-Mizil.
\newblock Model compression.
\newblock In \emph{KDD}, pages 535--541. ACM, 2006.

\bibitem[Caruana(1997)]{Caruana1997multitask}
Rich Caruana.
\newblock Multitask learning.
\newblock \emph{Mach. Learn.}, 28\penalty0 (1):\penalty0 41--75, July 1997.

\bibitem[Chan et~al.(2015)Chan, Ke, and Lane]{chan2015transferring}
William Chan, Nan~Rosemary Ke, and Ian Lane.
\newblock Transferring knowledge from a rnn to a dnn.
\newblock \emph{arXiv preprint arXiv:1504.01483}, 2015.

\bibitem[Farahmand et~al.(2012)Farahmand, Precup, and
  Ghavamzadeh]{farahmand2012}
Amir-massoud Farahmand, Doina Precup, and Mohammad Ghavamzadeh.
\newblock Generalized classification-based approximate policy iteration.
\newblock In \emph{Tenth European Workshop on Reinforcement Learning (EWRL)},
  volume~2, 2012.

\bibitem[Fleming and Wallace(1986)]{Fleming1986how}
Philip~J. Fleming and John~J. Wallace.
\newblock How not to lie with statistics: The correct way to summarize
  benchmark results.
\newblock \emph{Commun. ACM}, 29\penalty0 (3):\penalty0 218--221, March 1986.

\bibitem[Greensmith et~al.(2004)Greensmith, Bartlett, and
  Baxter]{Greensmith2004}
Evan Greensmith, Peter~L. Bartlett, and Jonathan Baxter.
\newblock Variance reduction techniques for gradient estimates in reinforcement
  learning.
\newblock \emph{Journal of Machine Learning Research (JMLR)}, pages 1471--1530,
  2004.

\bibitem[Guo et~al.(2014)Guo, Singh, Lee, Lewis, and Wang]{GuoSLLW2014}
Xiaoxiao Guo, Satinder~P. Singh, Honglak Lee, Richard~L. Lewis, and Xiaoshi
  Wang.
\newblock Deep learning for real-time atari game play using offline monte-carlo
  tree search planning.
\newblock In \emph{Advances in Neural Information Processing Systems (NIPS)},
  pages 3338--3346, 2014.

\bibitem[{Hinton} et~al.(2014){Hinton}, {Vinyals}, and
  {Dean}]{HintonDistillation2014}
G.~{Hinton}, O.~{Vinyals}, and J.~{Dean}.
\newblock {Distilling the Knowledge in a Neural Network}.
\newblock \emph{Deep Learning and Representation Learning Workshop, NIPS},
  2014.

\bibitem[Kocsis and Szepesv{\'a}ri(2006)]{kocsis2006bandit}
Levente Kocsis and Csaba Szepesv{\'a}ri.
\newblock Bandit based monte-carlo planning.
\newblock In \emph{Machine Learning: ECML 2006}, pages 282--293. Springer,
  2006.

\bibitem[Lazaric et~al.(2010)Lazaric, Ghavamzadeh, and Munos]{Lazaric2010}
Alessandro Lazaric, Mohammad Ghavamzadeh, and R{\'e}mi Munos.
\newblock Analysis of a classification-based policy iteration algorithm.
\newblock In \emph{ICML-27th International Conference on Machine Learning},
  pages 607--614. Omnipress, 2010.

\bibitem[Li et~al.(2014)Li, Zhao, Huang, and Gong]{li2014learning}
Jinyu Li, Rui Zhao, Jui-Ting Huang, and Yifan Gong.
\newblock Learning small-size dnn with output-distribution-based criteria.
\newblock In \emph{Proc. Interspeech}, 2014.

\bibitem[Liang et~al.(2008)Liang, III, and
  Klein]{LiangStructureCompilation2008}
Percy Liang, Hal~Daumé III, and Dan Klein.
\newblock Structure compilation: trading structure for features.
\newblock In \emph{Proceedings of International Conference on Machine Learning
  (ICML)}, 2008.

\bibitem[Menke and Martinez(2009)]{menke2009improving}
Joshua Menke and Tony Martinez.
\newblock Improving supervised learning by adapting the problem to the learner.
\newblock \emph{International Journal of Neural Systems}, 19\penalty0
  (01):\penalty0 1--9, 2009.

\bibitem[Mnih et~al.(2013)Mnih, Kavukcuoglu, Silver, Graves, Antonoglou,
  Wierstra, and Riedmiller]{MnihNips2013}
Volodymyr Mnih, Koray Kavukcuoglu, David Silver, Alex Graves, Ioannis
  Antonoglou, Daan Wierstra, and Martin~A. Riedmiller.
\newblock Playing atari with deep reinforcement learning.
\newblock \emph{Deep Learning Workshop, NIPS}, 2013.

\bibitem[Mnih et~al.(2015)Mnih, Kavukcuoglu, Silver, Rusu, Veness, Bellemare,
  Graves, Riedmiller, Fidjeland, Ostrovski, Petersen, Beattie, Sadik,
  Antonoglou, King, Kumaran, Wierstra, Legg, and Hassabis]{MnihNature2015}
Volodymyr Mnih, Koray Kavukcuoglu, David Silver, Andrei~A. Rusu, Joel Veness,
  Marc~G. Bellemare, Alex Graves, Martin Riedmiller, Andreas~K. Fidjeland,
  Georg Ostrovski, Stig Petersen, Charles Beattie, Amir Sadik, Ioannis
  Antonoglou, Helen King, Dharshan Kumaran, Daan Wierstra, Shane Legg, and
  Demis Hassabis.
\newblock Human-level control through deep reinforcement learning.
\newblock \emph{Nature}, 518\penalty0 (7540):\penalty0 529--533, 02 2015.

\bibitem[Nair et~al.(2015)Nair, Srinivasan, Blackwell, Alcicek, Fearon, Maria,
  Panneershelvam, Suleyman, Beattie, Petersen, Legg, Mnih, Kavukcuoglu, and
  Silver]{nair2015gorila}
Arun Nair, Praveen Srinivasan, Sam Blackwell, Cagdas Alcicek, Rory Fearon,
  Alessandro~De Maria, Vedavyas Panneershelvam, Mustafa Suleyman, Charles
  Beattie, Stig Petersen, Shane Legg, Volodymyr Mnih, Koray Kavukcuoglu, and
  David Silver.
\newblock Massively parallel methods for deep reinforcement learning.
\newblock \emph{CoRR}, abs/1507.04296, 2015.

\bibitem[Romero et~al.(2014)Romero, Ballas, Kahou, Chassang, Gatta, and
  Bengio]{romero2014fitnets}
Adriana Romero, Nicolas Ballas, Samira~Ebrahimi Kahou, Antoine Chassang, Carlo
  Gatta, and Yoshua Bengio.
\newblock Fitnets: Hints for thin deep nets.
\newblock \emph{arXiv preprint arXiv:1412.6550}, 2014.

\bibitem[Ross et~al.(2010)Ross, Gordon, and Bagnell]{ross2010}
St{\'e}phane Ross, Geoffrey~J Gordon, and J~Andrew Bagnell.
\newblock A reduction of imitation learning and structured prediction to
  no-regret online learning.
\newblock \emph{arXiv preprint arXiv:1011.0686}, 2010.

\bibitem[{Shalev-Shwartz}(2014)]{Shalev-Shwartz2014}
S.~{Shalev-Shwartz}.
\newblock {SelfieBoost: A Boosting Algorithm for Deep Learning}.
\newblock \emph{ArXiv e-prints}, November 2014.

\bibitem[Sutton and Barto(1998)]{SuttonAndBarto1998intro}
Richard~S. Sutton and Andrew~G. Barto.
\newblock \emph{Introduction to Reinforcement Learning}.
\newblock MIT Press, Cambridge, MA, USA, 1st edition, 1998.

\bibitem[Tang et~al.(2015)Tang, Wang, Pan, and Zhang]{tang2015knowledge}
Zhiyuan Tang, Dong Wang, Yiqiao Pan, and Zhiyong Zhang.
\newblock Knowledge transfer pre-training.
\newblock \emph{arXiv preprint arXiv:1506.02256}, 2015.

\bibitem[Tieleman and Hinton(2012)]{Tieleman2012rmsprop}
T.~Tieleman and G.~Hinton.
\newblock {Lecture 6.5---RmsProp: Divide the gradient by a running average of
  its recent magnitude}.
\newblock COURSERA: Neural Networks for Machine Learning, 2012.

\bibitem[van~der Maaten and Hinton(2008)]{maaten2008visualizing}
L.J.P. van~der Maaten and G.E. Hinton.
\newblock Visualizing high-dimensional data using t-sne.
\newblock \emph{Journal of Machine Learning Research (JMLR)}, 2008.

\bibitem[van Hasselt et~al.()van Hasselt, Guez, and
  Silver]{vanHasselt2015double}
Hado van Hasselt, Arthur Guez, and David Silver.
\newblock Deep reinforcement learning with double q-learning.
\newblock In \emph{Proceedings of the AAAI Conference on Artificial
  Intelligence}.

\bibitem[Wang et~al.(2015)Wang, Liu, Tang, Zhang, and Zhao]{wang2015recurrent}
Dong Wang, Chao Liu, Zhiyuan Tang, Zhiyong Zhang, and Mengyuan Zhao.
\newblock Recurrent neural network training with dark knowledge transfer.
\newblock \emph{arXiv preprint arXiv:1505.04630}, 2015.

\end{thebibliography}
\bibliographystyle{plainnat}

\appendix
\newpage

\setcounter{table}{0}
\setcounter{figure}{0}
\renewcommand{\thetable}{\thesection\arabic{table}}
\renewcommand{\thefigure}{\thesection\arabic{figure}}

\section{Experimental Details}
\label{AppendixA}

\textbf{Policy Distillation Training Procedure}
Online data collection during policy distillation was performed under similar conditions to agent evaluation in \cite{MnihNature2015}. The DQN agent plays a random number of null-ops (up to 30) to initialize the episode, then acts greedily with respect to its Q-function, except for 5\% of actions, which are chosen uniformly at random. Episodes can last up to 30 minutes of real-time play, or 108,000 frames. The small percentage of random actions leads to diverse game trajectories, which improves coverage of a game's state space.

We recorded the DQN teacher's outputs (Q-values for all valid actions) and inputs (emulator frames) into a \emph{replay memory} with a capacity of 10 hours of real-time gameplay (540,000 control steps at 15Hz). At the end of each new hour of teacher gameplay added to the replay memory we performed 10,000 minibatch updates on the student network. We used the RmsProp \citep{Tieleman2012rmsprop}  variation of minibatch stochastic gradient descent to train student networks. Results were robust for primary learning rates between $1.0e^{-4}$ and $1.0e^{-3}$, with maximum learning rates between $1.0e^{-3}$ and $1.0e^{-1}$. We chose hyper-parameters using preliminary experiments on 4 games. The reported results consumed 500 hours of teacher gameplay to train each student, less than 50\% of the amount that was used to train each DQN teacher. Using modern GPUs we can refresh the replay memory and train the students much faster than real-time, with typical convergence in a few days. With multi-task students we used separate replay memories for each game, with the same capacity of 10 hours, and the respective DQN teachers took turns adding data. After one hour of gameplay the student is trained with 10,000 minibatch updates (each minibatch is drawn from a randomly chosen single game memory). The same 500 hour budget of gameplay was used for all but the largest network, which used 34,000 hours of experience over 10 games.


\textbf{Distillation Targets}
Using DQN outputs we have defined three types of training targets that correspond to the three distillation loss functions discussed in Section\ref{sec-approach}. First, the teacher's Q-values for all actions were used directly as supervised targets; thus, training the student consisted of minimizing the mean squared error (MSE) between the student's and teacher's outputs for each input state.
Second, we used only the teacher's highest valued action as a one-hot target. Naturally, we minimized the negative log likelihood (NLL) loss.  Finally, we passed Q-values through a softmax function whose temperature $(\tau = 0.01)$ was selected empirically from $[1.0, 0.1, 0.01, 0.001]$. The resulting probabilities were used as targets by minimizing the Kullback-Leibler (KL) divergence between these ``sharpened'' action probabilities and the corresponding output distribution predicted by the student policy. We went on to use the KL cost function for a majority of reported experiments, with a fixed hyper-parameter value. This choice was based on experiments described in subsection \ref{comparison_of_cost_functions_subsec} which were performed on 4 out of the 10 games considered.

\textbf{Network Architectures}
Details of the architectures used by DQN and single-task distilled agents are given in table \ref{network_size_table}. Rectifier non-linearities were added between each two consecutive layers. We used one unit for each valid action in the output layer, which was linear. A final softmax operation was performed when distilling with NLL and KL loss functions.

\begin{table}[h]
\centering
\caption{Network architectures and parameter counts of models used for single-task compression experiments.}
\scalebox{0.8}{
\begin{tabular}{l|c|c|c|c|c|c|c}
\toprule
Agent           & Input  & Conv. 1  & Conv. 2  & Conv. 3  & F.C. 1 & Output    & Parameters
\\ \midrule
Teacher (DQN)   & 4      & 32       & 64       & 64       & 512    & up to 18  & 1,693,362
\\ \midrule
Dist-KL-net1    & 4      & 16       & 32       & 32       & 256    & up to 18  & 427,874
\\ \midrule
Dist-KL-net2    & 4      & 16       & 16       & 16       & 128    & up to 18  & 113,346
\\ \midrule
Dist-KL-net3    & 4      & 16       & 16       & 16       & 64     & up to 18  & 61,954
\\ \bottomrule
\end{tabular}
}
\label{network_size_table}
\end{table}

For compression experiments we scaled down the number of units in each layer without changing the basic architecture. The vast majority of saved parameters were in the fully connected layer on top of the convolutional stack.

The distinct characteristic of all multi-task experiments was the use of different MLP ``controller'' networks for each game, on top of shared representations. The specific details of these architectures are given in table \ref{network_size_table_multi}. All results reported on 3 games used identical models of similar size with a single DQN teacher. A network 4 times larger than a teacher was trained using multi-task distillation on 10 games.

\begin{table}[h]
\centering
\caption{Network architectures and parameter counts of models used for multi-task distillation experiments.}
\scalebox{0.8}{
\begin{tabular}{l|c|c|c|c|c|c|c|c}
\toprule
Agent                     & Input & Conv. 1 & Conv. 2 & Conv. 3 & F.C. 1 & F.C. 2   & Output         & Parameters
\\ \midrule
One Teacher (DQN)         & 4     & 32      & 64      & 64      & 512    & n/a      & up to 18       & 1,693,362
\\ \midrule
Multi-DQN/Dist (3 games)  & 4     & 32      & 64      & 64      & 512    & 128 (x3) & up to 18 (x3)  & 1,882,668
\\ \midrule
Multi-Dist-KL (10 games)  & 4     & 64      & 64      & 64      & 1500   & 128 (x10)& up to 18 (x10) & 6,756,721
\\ \bottomrule
\end{tabular}
}
\label{network_size_table_multi}
\end{table}

\textbf{Agent Evaluation}
Professional human expert play was used to generate starting states for each game by sampling 100 random positions which occurred in the first 20\% of each episode's length. Agents are allowed to act for 30 minutes of real-time gameplay, or 108,000 frames, and they use a high value of $\epsilon$ equal to $5\%$. We also do not compute a generalization score until the agent's training process has ended.

Evaluating the performance of a multi-task agent is not trivial.
Since each game has a different reward structure and somewhat arbitrary choice of reward scale, it is meaningless to compute an arithmetic mean of scores across games. Therefore, DQN generalization scores (published previously \citep{vanHasselt2015double, nair2015gorila}) are taken as a reference point and student scores are reported as a relative percentage.
This way, performance on multiple games can be measured using the geometric mean \citep{Fleming1986how}.



\setcounter{table}{0}
\setcounter{figure}{0}
\renewcommand{\thetable}{\thesection\arabic{table}}
\renewcommand{\thefigure}{\thesection\arabic{figure}}
\section{Supporting Tables for Policy Distillation Figures}
\label{AppendixB}

\begin{table}[h]
\centering
\caption{Performance of single-task compressed networks on 10 Atari games. Best relative scores are outlined in bold.}
\label{compression-full-table}
\scalebox{0.9}{
\begin{tabular}{@{}l|l|ll|ll|ll@{}}
\toprule
\textbf{}               & DQN             & \multicolumn{2}{l}{Dist-KL-net1} & \multicolumn{2}{l}{Dist-KL-net2} & \multicolumn{2}{l}{Dist-KL-net3} \\ \midrule
                        & \textbf{score} & \textbf{score}  & \textbf{\% DQN} & \textbf{score}  & \textbf{\% DQN}   & \textbf{score}  & \textbf{\% DQN}  \\
\textbf{Beamrider}      & 8672.4          & 7552.8    &  \textbf{87.1}       &  7393.3     &  85.3              &  6521.2     & 75.2
\\
\textbf{Breakout}       & 303.9           & 321.0     &  \textbf{105.6}      &  298.2      &  98.1              &  238.8      & 78.6
\\
\textbf{Enduro}         & 475.6           & 677.9     &  \textbf{142.5}      &  672.2      &  141.3             &  556.7      & 117.1
\\
\textbf{Freeway}        & 25.8            & 26.7      &  \textbf{103.5}      &  26.7       &  \textbf{103.5}    &  26.7       & \textbf{103.5}
\\
\textbf{Ms.Pacman}      & 763.5           & 782.5     &  \textbf{102.5}      &  659.9      &  86.4              &  734.3      & 96.2
\\
\textbf{Pong}           & 16.2            & 16.3      &  100.6               &  16.8       &  \textbf{103.7}    &  15.7       & 96.9
\\
\textbf{Q*bert}         & 4589.8          & 5947.3    &  129.6               &  5994.0     &  \textbf{130.6}    &  4952.3     & 107.9
\\
\textbf{Riverraid}      & 4065.3          & 4442.7    &  \textbf{109.3}      &  4175.3     &  102.7             &  3417.9     & 84.1
\\
\textbf{Seaquest}       & 2793.3          & 3986.6    &  142.7               &  4567.1     &  \textbf{163.5}    &  3838.3     & 137.4
\\
\textbf{Space Invaders} & 1449.7          & 1140.0    &  \textbf{78.6}       &  716.1      &  49.4              &  302.3      & 20.9
\\  \midrule
\textbf{Geometric Mean} &                 &           &  \textbf{108.3}      &             &  101.7             &             & 83.9
\\ \bottomrule
\end{tabular}
}
\end{table}

\begin{table}[h]
\centering
\caption{Performance of multi-task distilled agents on 3 Atari games. Best relative scores are outlined in bold.}
\label{atari3-full-table}
\scalebox{0.9}{
\begin{tabular}{@{}l|l|ll|ll|ll@{}}
\toprule
\textbf{}               & DQN  & \multicolumn{2}{l}{Multi-DQN}  & \multicolumn{2}{l}{Multi-Dist-NLL}   & \multicolumn{2}{l}{Multi-Dist-KL} \\ \midrule
                        & \textbf{score} & \textbf{score}  & \textbf{\% DQN} & \textbf{score}  & \textbf{\% DQN}   & \textbf{score}  & \textbf{\% DQN}  \\
\textbf{Freeway}        & 25.8  & 23.3   & 90.3 & 26.5     & \textbf{102.7} & 26.3   & \textbf{102.0}
\\
\textbf{Pong}           & 16.2  & 12.0   & 74.1 & 14.8     & 91.4  & 16.9   & \textbf{104.4}
\\
\textbf{Q*bert}         & 4589.8& 3987.3 & 86.9 & 5678.0   & 123.7 & 6890.3 & \textbf{150.1}
\\  \midrule
\textbf{Geometric Mean} &       &        & 83.5 &          & 105.1 &        & \textbf{116.9}
\\ \bottomrule
\end{tabular}
}
\end{table}

\newpage

\setcounter{table}{0}
\setcounter{figure}{0}
\renewcommand{\thetable}{\thesection\arabic{table}}
\renewcommand{\thefigure}{\thesection\arabic{figure}}

\section{Additional Results Using Online Policy Distillation}
\label{AppendixC}

\begin{figure*}[h]
  \centering
    \includegraphics[width=0.80\textwidth]{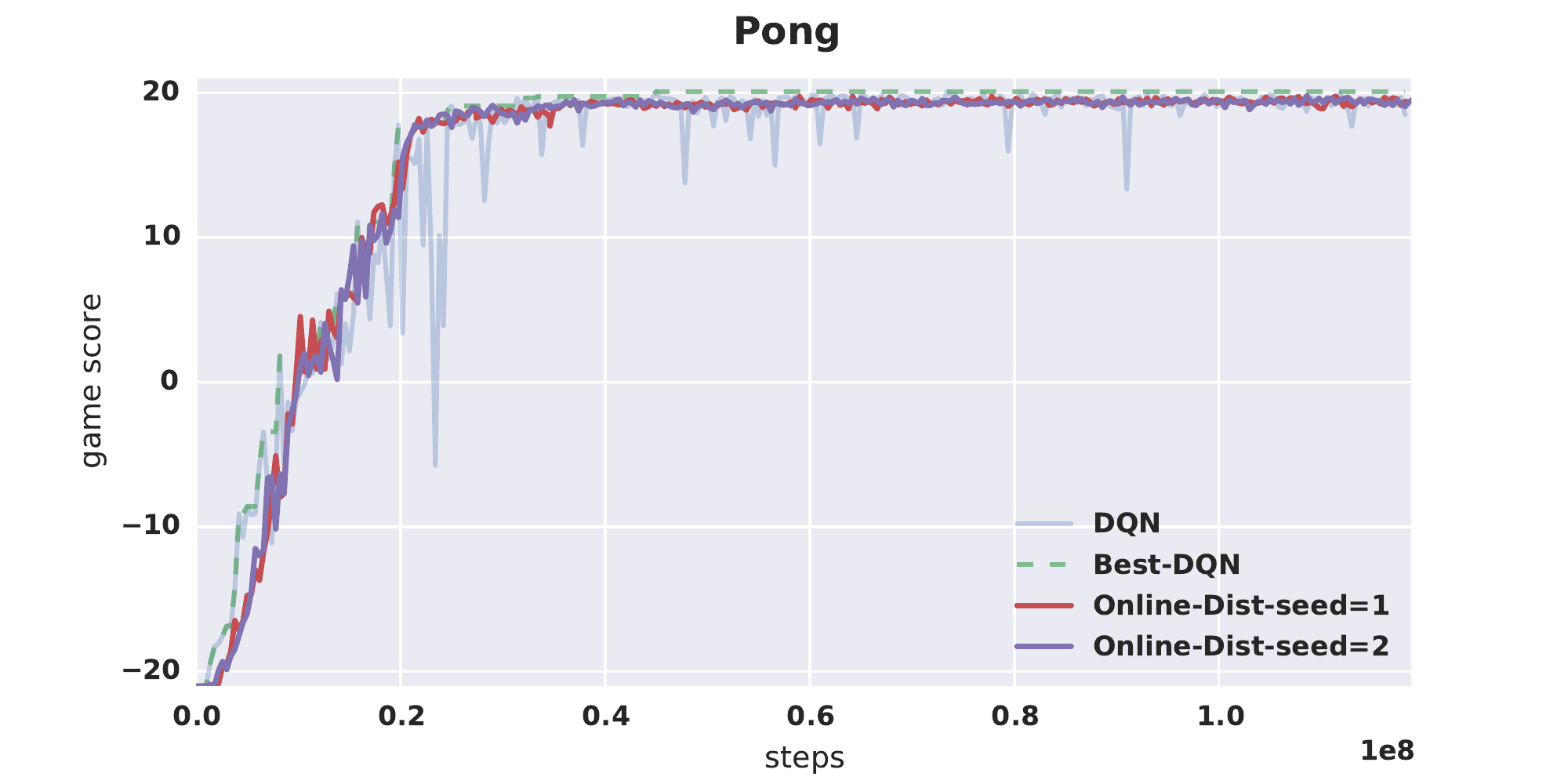}\\
    \includegraphics[width=0.80\textwidth]{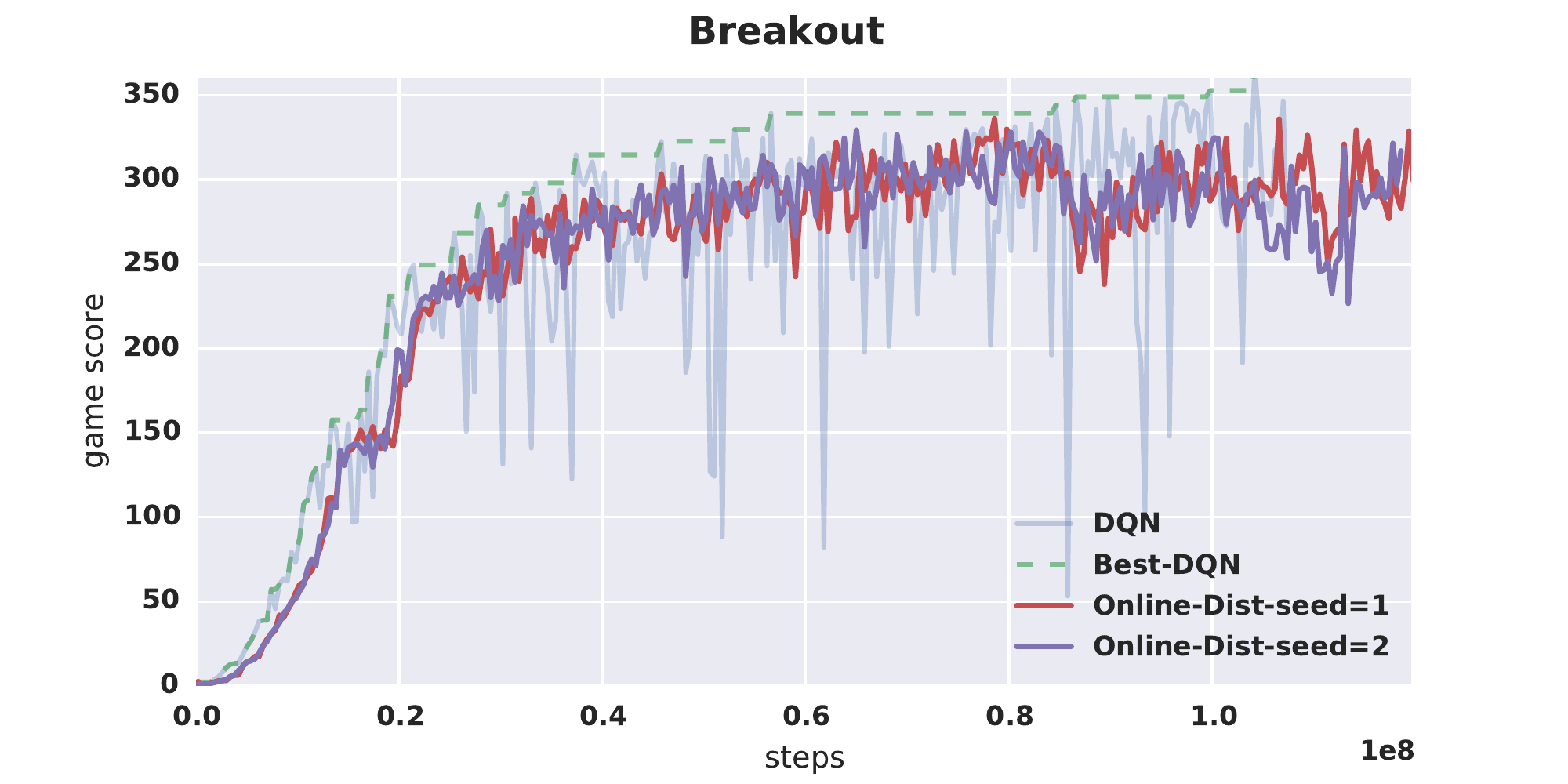}\\
    \caption{Online Policy Distillation during DQN learning (pale blue). The current best DQN policy to date (green) is distilled into a new network  during DQN training. Showing online distillation experiments on 2 games, with 2 initial random seeds and the same learning rate across all runs.}
  \label{online_dist_fig_supl}
\end{figure*}

\section{Visualization of representation over 10 Atari games}
\label{AppendixD}

\begin{figure*}[h]
  \centering
    \includegraphics[width=0.75\textwidth]{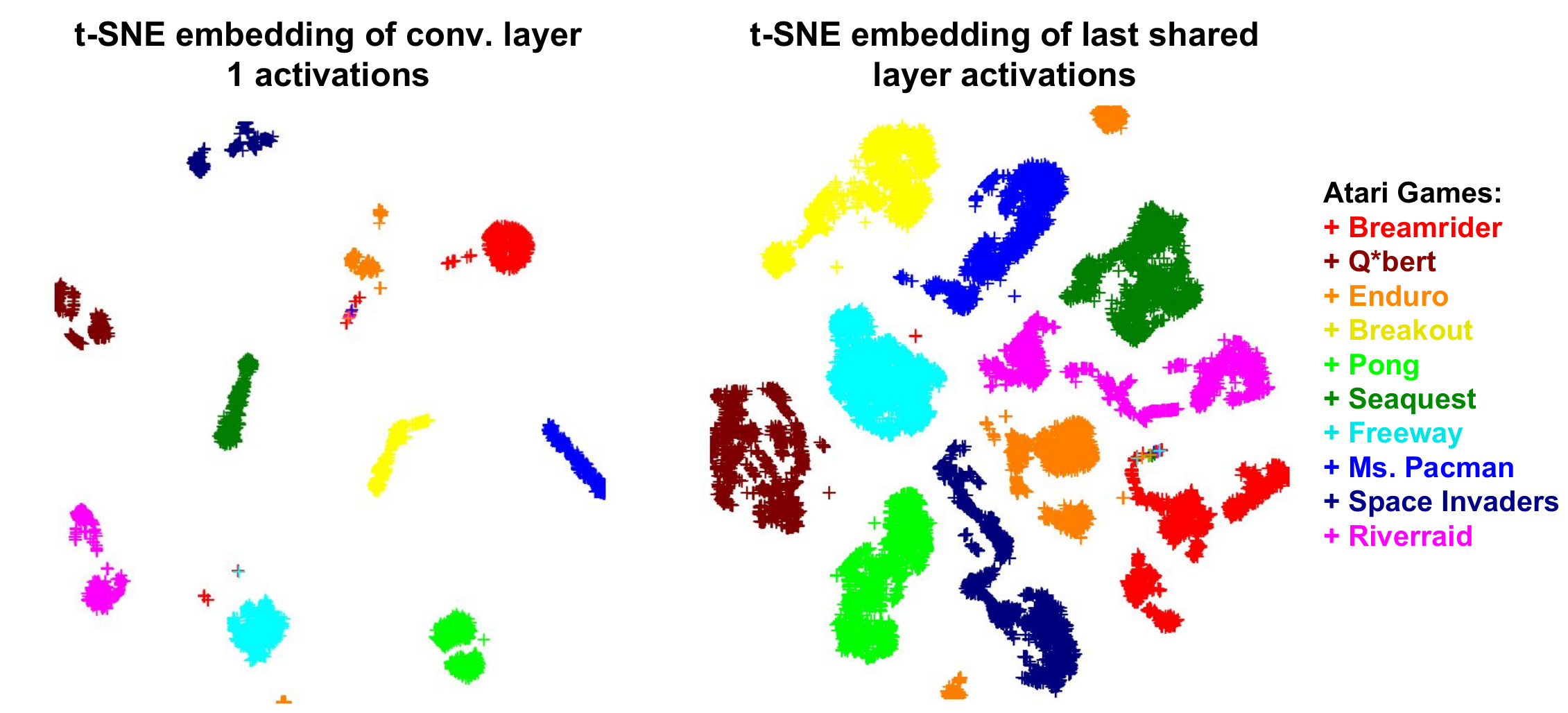}
    \caption{Shared embeddings learned using Multi-Dist-KL on data from 10 Atari games. Showing t-SNE visualizations \citep{maaten2008visualizing} of activations in the first convolutional and last fully connected layers.}
  \label{activations_fig}
\end{figure*}

A visual interpretation of the representation learned by the multi-task distillation  is given in Figure \ref{activations_fig}, where t-SNE embeddings of network activations from 10 different games are plotted with distinct colors. The embedding on the left suggests that statistics of low level representations (layer 1) may be game-specific, although this may be due to the diversity of the inputs. The second embedding characterizes shared representations at the final network layer. While activations are still game-specific, we observe higher within-game variance of representations, which probably reflect output statistics.

\end{document}